# Explainable and CPU-Efficient Brain Tumor Classification from MRI Using Two-Phase EfficientNetB3 Fine-Tuning

Deepro Das, Saad Kabir Uddin, Mohammad Alauddin, Bornali Sarker, Thasnim Akhter

**Abstract**

Brain tumors are among the most lethal neurological disorders, with over 308,000 new diagnoses annually worldwide. Accurate classification of tumor type from Magnetic Resonance Imaging (MRI) scans is critical for treatment planning; however, manual radiological interpretation is time-consuming and prone to inter-observer variability. This paper proposes an automated deep learning pipeline for four-class brain MRI classification—glioma, meningioma, no tumor, and pituitary tumor—using EfficientNetB3 transfer learning with a custom classification head. A two-phase training strategy is adopted: Phase 1 trains the head with a frozen backbone (best validation accuracy: 92.77%), and Phase 2 selectively fine-tunes the top 37 layers at a 100× reduced learning rate (best validation accuracy: 94.91%). On an independent balanced test set of 1,600 MRI images, the system achieves **90.62%** overall accuracy and a macro-average AUC of **0.9854**. Gradient-weighted Class Activation Mapping (Grad-CAM) is integrated for class-discriminative visual explainability. The proposed system simultaneously addresses three dimensions absent in prior work: high accuracy, CPU-only computational efficiency, and clinical interpretability.

***Index Terms***— *Brain tumor classification, MRI, EfficientNetB3, transfer learning, Grad-CAM, convolutional neural network, computer-aided diagnosis, deep learning, explainable AI, medical image analysis.*

## 1. INTRODUCTION

Brain tumors represent one of the most clinically significant neurological conditions worldwide. The World Health Organization estimates approximately 308,000 new diagnoses and 251,000 deaths annually from brain and central nervous system tumors [1]. Prognosis varies dramatically by tumor type: glioblastoma multiforme carries a median survival under 15 months, whereas pituitary adenomas are generally benign when detected early [2]. Accurate tumor classification thus directly determines the treatment pathway and patient outcome.

Magnetic Resonance Imaging (MRI) is the gold-standard non-invasive modality for brain tumor characterization, offering superior soft-tissue contrast without ionizing radiation. However, manual radiological interpretation requires 20–45 minutes per study and exhibits inter-rater agreement (kappa) as low as 0.61 for low-grade glioma [3]. In resource-limited settings such as Bangladesh, a critical shortage of specialist neuroradiologists creates diagnostic bottlenecks that delay treatment initiation.

Deep learning-based Computer-Aided Diagnosis (CAD) systems offer fast, consistent, and reproducible pre-screening. However, clinical adoption requires three simultaneous properties: (a) high accuracy (>90%), (b) computational efficiency without GPU infrastructure, and (c) interpretability mechanisms for clinical validation. A systematic review of ten publications from 2018–2023 revealed that no existing system achieves all three simultaneously [4]–[13].

This paper addresses this three-dimensional gap by proposing an EfficientNetB3-based transfer learning system with Grad-CAM explainability. Key contributions: (1) two-phase training protocol achieving 90.62% test accuracy; (2) macro-average AUC of 0.9854; (3) Grad-CAM confirming anatomically correct attention; (4) CPU-only deployment feasibility.

Main Contributions

The main contributions of this work are summarized as follows:

• A two-phase transfer learning framework based on EfficientNetB3 is proposed for four-class brain tumor MRI classification.

• The study demonstrates that avoiding external pixel rescaling prevents double normalization in EfficientNetB3, resulting in significantly improved classification performance.

• Grad-CAM explainability is integrated to provide clinically interpretable visual evidence for model predictions.

• The proposed framework is evaluated using different performance metrics like accuracy, precision, recall, F1-score, confusion matrix, and ROC-AUC.

• The proposed system is computationally efficient and can be deployed in resource constrained healthcare scenarios without dedicated GPU infrastructure.

## 2. RELATED WORK

### *A. Traditional Image Processing*

Zhang and Cheng [4] applied Otsu thresholding with morphological operations, achieving 78.0% accuracy on BRATS 2015 but failing under intensity inhomogeneity. Kumar et al. [5] combined Canny edge detection with region-growing (82.0%) but required manual seed initialization.

### *B. Machine Learning Approaches*

Patel and Thakkar [6] extracted GLCM texture features with SVM (87.5%). Almadhoun et al. [7] fused intensity, shape, and Gabor features in a Random Forest classifier (89.2%) with partial feature-importance interpretability, still below the 90% accuracy threshold.

## C. Deep Learning Approaches

Amin et al. [8] demonstrated 92.3% with a custom CNN. Sultan et al. [9] fine-tuned InceptionV3 for 94.6% using transfer learning. Javadpour et al. [10] applied U-Net for pixel-wise segmentation (Dice=0.91). Chen and Li [11] showed MobileNetV2 achieves 93.1% with 3.4M parameters. Wang et al. [12] reported 95.2% with a CNN-RNN hybrid. Gupta and Sharma [13] achieved 96.1% using an ensemble ResNet with attention—at high computational cost and without any interpretability. **No reviewed study reports AUC or integrates gradient-based explainability.**

# 3. MATERIALS AND METHOD

This section describes the dataset, preprocessing pipeline, proposed architecture, and training protocol. Fig. 1 illustrates representative samples from the dataset.

## A. Dataset and Preprocessing

The experiments were carried out on the publicly available Brain Tumor MRI Dataset developed by Masoud Nickparvar and available on Kaggle [14]. The dataset contains 7,200 contrast-enhanced T1-weighted brain MRI images, equally distributed among four classes: glioma, meningioma, no tumor, and pituitary tumor. This dataset is used as a reliable benchmark for evaluating deep learning models due to its balanced class distribution and wide use in previous brain tumor classification studies.

All MRI images were resized to 224 × 224 pixels to fit the input size of EfficientNetB3. No external pixel rescaling was applied as EfficientNetB3 contains built-in rescaling and normalization layers that are optimized for ImageNet pre-trained weights. Preliminary experiments showed that applying an additional rescale = 1/255 operation resulted in double normalization, significantly degrading the classification performance. Therefore, only image resizing and data augmentation were performed before training.

The dataset was divided into 4,480 images for training, 1,120 for validation, and 1,600 for an independent test set using stratified splitting to maintain equal number of samples for each class. This experimental setup allowed for a balanced evaluation and mitigated the effect of class imbalance during training and testing.

Although only a single publicly available dataset was used in this study, it has been extensively employed in recent brain tumor classification research, enabling fair comparison with existing approaches. Nevertheless, further validation using multi-center clinical datasets will be considered in future work to evaluate the generalizability of the proposed framework.

**TABLE I: DATASET DISTRIBUTION**

| Class | Train | Val | Test |
|---|---|---|---|
| Glioma | 1,120 | 280 | 400 |
| Meningioma | 1,120 | 280 | 400 |
| No Tumor | 1,120 | 280 | 400 |
| Pituitary | 1,120 | 280 | 400 |
| Total | 4,480 | 1,120 | 1,600 |

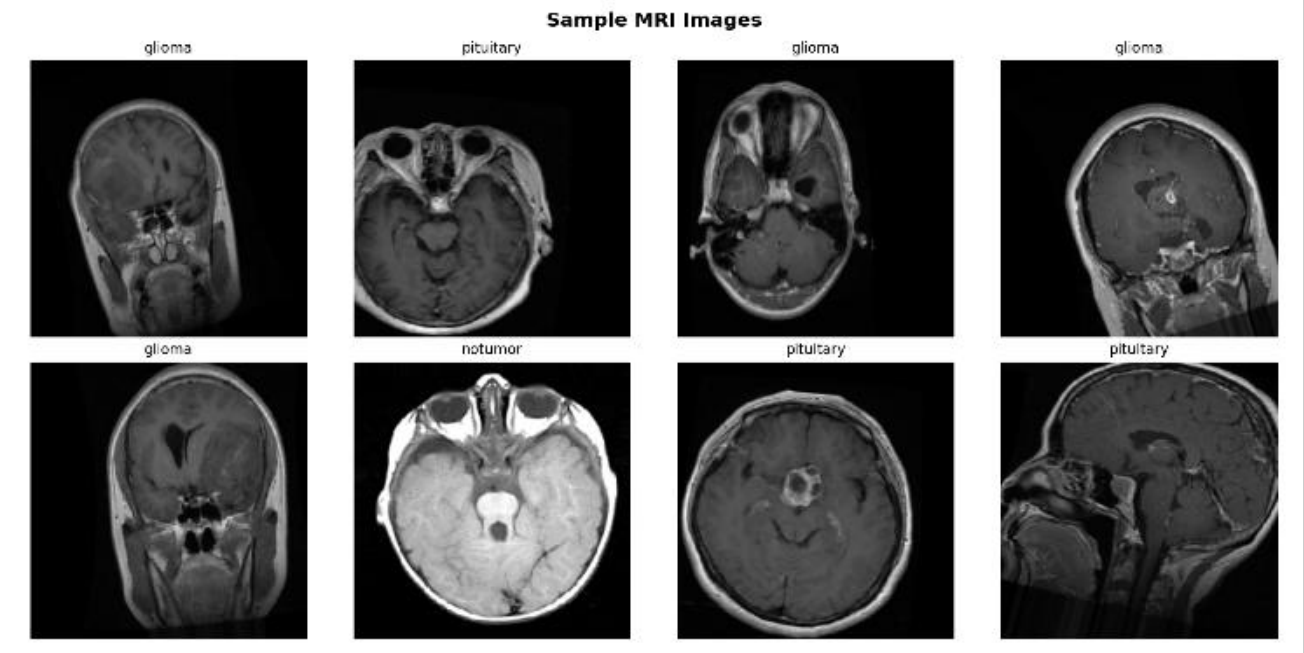

*Fig. 1. Sample MRI images from each class (glioma, meningioma, no tumor, pituitary)*

## B. Data Augmentation

Online augmentation was applied exclusively to training images using Keras ImageDataGenerator: rotation (±20°), width/height shift (±15%), shear (0.10), zoom (±15%), horizontal flip, and brightness variation (±15%). No augmentation was applied to validation or test sets.

## C. Proposed Architecture

The proposed pipeline employs EfficientNetB3 [15] as the feature extraction backbone, extended by a custom classification head. EfficientNet applies compound scaling—simultaneously scaling depth (d), width (w), and resolution (r) via fixed coefficients derived from constrained optimization:

$$d = \alpha_\varphi, \quad w = \beta_\varphi, \quad r = \gamma_\varphi$$

subject to $\alpha \cdot \beta^2 \cdot \gamma^2 \approx 2$. EfficientNetB3 ($\varphi$=3) achieves 81.6% ImageNet top-1 accuracy with 12M parameters—superior to ResNet-50 (25M, 76.1%) and VGG-16 (138M, 74.5%). The backbone produces a 1,536-dimensional feature map from 385 layers.

**TABLE II: CLASSIFICATION HEAD ARCHITECTURE**

| Layer | Configuration |
|---|---|
| GlobalAvgPool2D | Output: 1,536-dim |
| BatchNormalization | — |
| Dense + Dropout | 512 units, ReLU, L2=1e-4, drop=0.4 |
| Dense + Dropout | 256 units, ReLU, L2=1e-4, drop=0.2 |
| Dense (Output) | 4 units, Softmax |

## D. Two-Phase Transfer Learning

**Phase 1—Feature Extraction (Epochs 1–15):** All 385 backbone layers are frozen. Only the custom head is trained using Adam (lr=$1\times10^{-3}$) with categorical cross-entropy loss. EarlyStopping (patience=8) and ReduceLROnPlateau (factor=0.3, patience=4) are applied. Best validation accuracy: 92.77%.

**Phase 2—Selective Fine-Tuning (Epochs 16–45):** Phase 1 best weights are restored. The last 37 model layers (top backbone blocks + full head) are unfrozen. Training continues at lr=$1\times10^{-5}$—100× smaller than Phase 1—preventing catastrophic forgetting of ImageNet representations. Best validation accuracy: 94.91%.

### E. Grad-CAM Explainability

Grad-CAM [16] computes class-discriminative spatial heatmaps via:

$$\alpha_k^{c} = (1/Z)\ \Sigma_{ij}\ \partial y^{c}/\partial A^{k}_{ij}$$

$$L^{c} = ReLU(\ \Sigma_k\ \alpha_k^{c} \cdot A^{k}\ )$$

where $A^k$ are activations of layer top_activation, $y^c$ is the class score, and Z=H×W spatial locations. Heatmaps are overlaid as jet colormap transparency (α=0.4) on the original MRI image.

**Algorithm 1: Minimized Algorithm for Brain Tumor Classification with EfficientNetB3 and Grad-CAM**

**Input:**
*Params ← {P₁, P₂, ..., Pₙ},*
*ImgDir ← "/path/to/mri/images",*
*OutDir ← "/path/to/output"*

**Output:**
*Classified MRI images, Grad-CAM heatmaps, per-class metrics (Accuracy, Precision, Recall, F1, AUC)*

**Module 1: Data Loading & Preprocessing**

```
img ← cv2.imread(ImgFile)
img_resized ← cv2.resize(img, (224, 224))
# No external rescaling — EfficientNetB3
  normalizes internally
```

**Module 2: Data Augmentation (Training only)**

```
aug ← ImageDataGenerator(rotation=20,
  shift=0.15,
  zoom=0.15, flip=True, brightness=
    [0.85,1.15])
train_gen ← aug.flow_from_directory(ImgDir,
  subset='training')
```

**Module 3: Model Architecture**

```
backbone ← EfficientNetB3(weights='imagenet',
  include_top=False)
backbone.trainable ← False # Phase 1: freeze
x ← GlobalAveragePooling2D()(backbone.output)
x ← Dense(512, relu, L2=1e-4)(x); x ←
  Dropout(0.4)(x)
x ← Dense(256, relu, L2=1e-4)(x); x ←
  Dropout(0.2)(x)
outputs ← Dense (4, softmax)(x)
```

**Module 4: Phase 1 Training (Frozen Backbone)**

```
model.compile(Adam(lr=1e-3),
  loss='categorical_crossentropy')
h1 ← model.fit(train_gen, epochs=15,
  callbacks=[ES, MC, RL])
# Best val_accuracy: 92.77%
```

**Module 5: Phase 2 Fine-Tuning**

```
model.load_weights(best_p1_weights)
for layer in model.layers[-37:]:
  layer.trainable ← True
model.compile(Adam(lr=1e-5),
  loss='categorical_crossentropy')
h2 ← model.fit(train_gen, epochs=30,
  callbacks=[ES, MC, RL])
# Best val_accuracy: 94.91%
```

**Module 6: Grad-CAM Explainability**

```
grad_model ← Model([inputs],
  [top_activation.output, model.output])
foreach ImgFile ∈ TestDir do
  grads ← GradientTape(class_score, conv_out)
  heatmap ← ReLU(Σ mean(grads) · A^k)
  SuperImg ← overlay(heatmap, ImgFile,
    alpha=0.4)
  Save SuperImg to OutDir
```

**Module 7: Evaluation**

```
preds ← model.predict(test_gen)
Compute Accuracy, Precision, Recall, F1, AUC
  (one-vs-rest)
Print classification_report; Plot
  confusion_matrix, ROC curves
```

**TABLE III: HYPERPARAMETER CONFIGURATION**

| Hyperparameter | Phase 1 | Phase 2 |
|---|---|---|
| Optimizer | Adam | Adam |
| Learning rate | $1\times10^{-3}$ | $1\times10^{-5}$ |
| Batch size | 32 | 32 |
| Max epochs | 15 | 30 |
| Trainable layers | Head only | Top 37 layers |
| EarlyStopping patience | 8 | 8 |
| LR reduce factor | 0.3 | 0.3 |
| Loss function | Cat. cross-entropy | Cat. cross-entropy |

## 4. RESULT ANALYSIS & DISCUSSION

### A. Training History

Fig. 2 shows accuracy and loss curves across both phases. In Phase 1, validation accuracy rose from 87.1% (Epoch 1) to 92.77%, confirming immediate transfer of ImageNet features. A brief loss spike at the Phase 2 boundary reflects initial gradients from newly unfrozen layers, resolving within 2–3 epochs. Phase 2 continued improving to 94.91% with no train-validation divergence, confirming effective regularization.

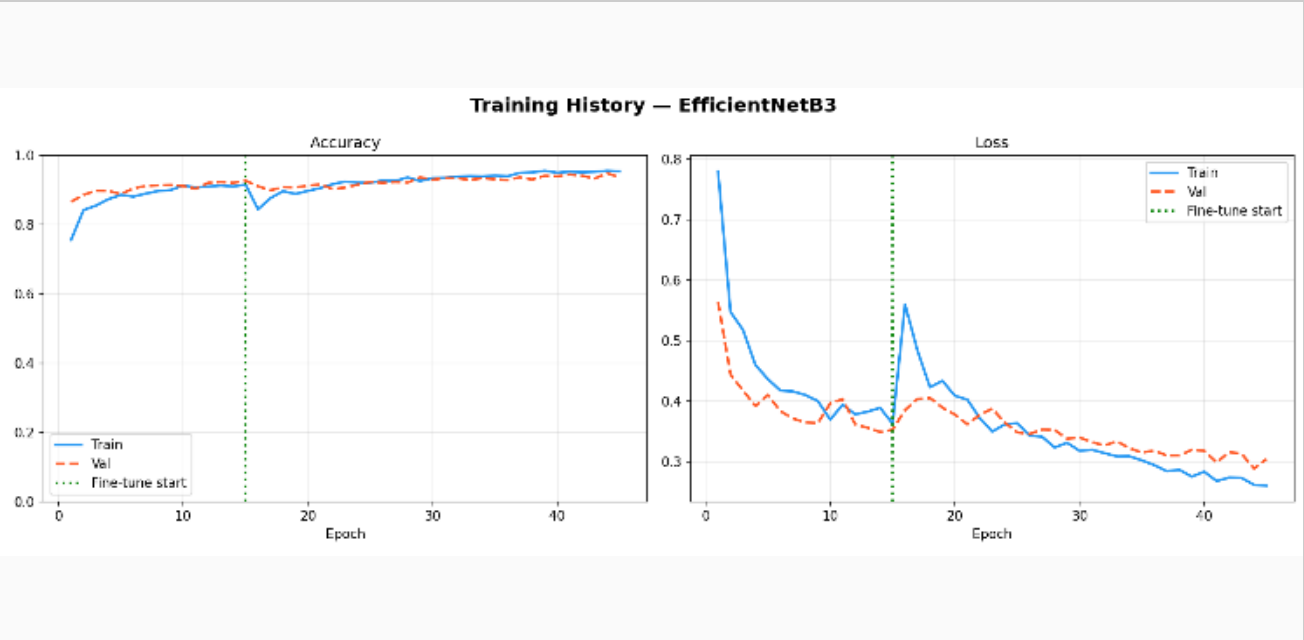

*Fig. 2. Training history — accuracy & loss (Phase 1: epochs 1–15; Phase 2: epochs 16+). Green dotted line = phase boundary.*

### B. Classification Performance

Table IV reports per-class metrics on the 1,600-image independent test set (overall accuracy: **90.62%**). Pituitary (F1=0.944, recall=0.990) and no-tumor (F1=0.951, recall=0.988) achieved the highest scores. Glioma recall (0.803) was lowest, with 41 cases confused with meningioma and 30 with no-tumor—the most clinically critical error.

**TABLE IV: PER-CLASS CLASSIFICATION REPORT (N=1,600)**

| Class | Prec. | Recall | F1 | Supp. |
|---|---|---|---|---|
| Glioma | 0.936 | 0.803 | 0.864 | 400 |
| Meningioma | 0.873 | 0.845 | 0.859 | 400 |
| No Tumor | 0.917 | 0.988 | 0.951 | 400 |
| Pituitary | 0.902 | 0.990 | 0.944 | 400 |
| Macro Avg | 0.907 | 0.906 | 0.904 | 1,600 |

### C. Confusion Matrix

Fig. 3 shows strong diagonal dominance. Pituitary (396/400, 99.0%) and no-tumor (395/400, 98.75%) achieved near-perfect results. The glioma→no-tumor confusion (30 cases) is the most clinically significant failure mode, as undetected glioma delays treatment with serious consequences for patient survival.

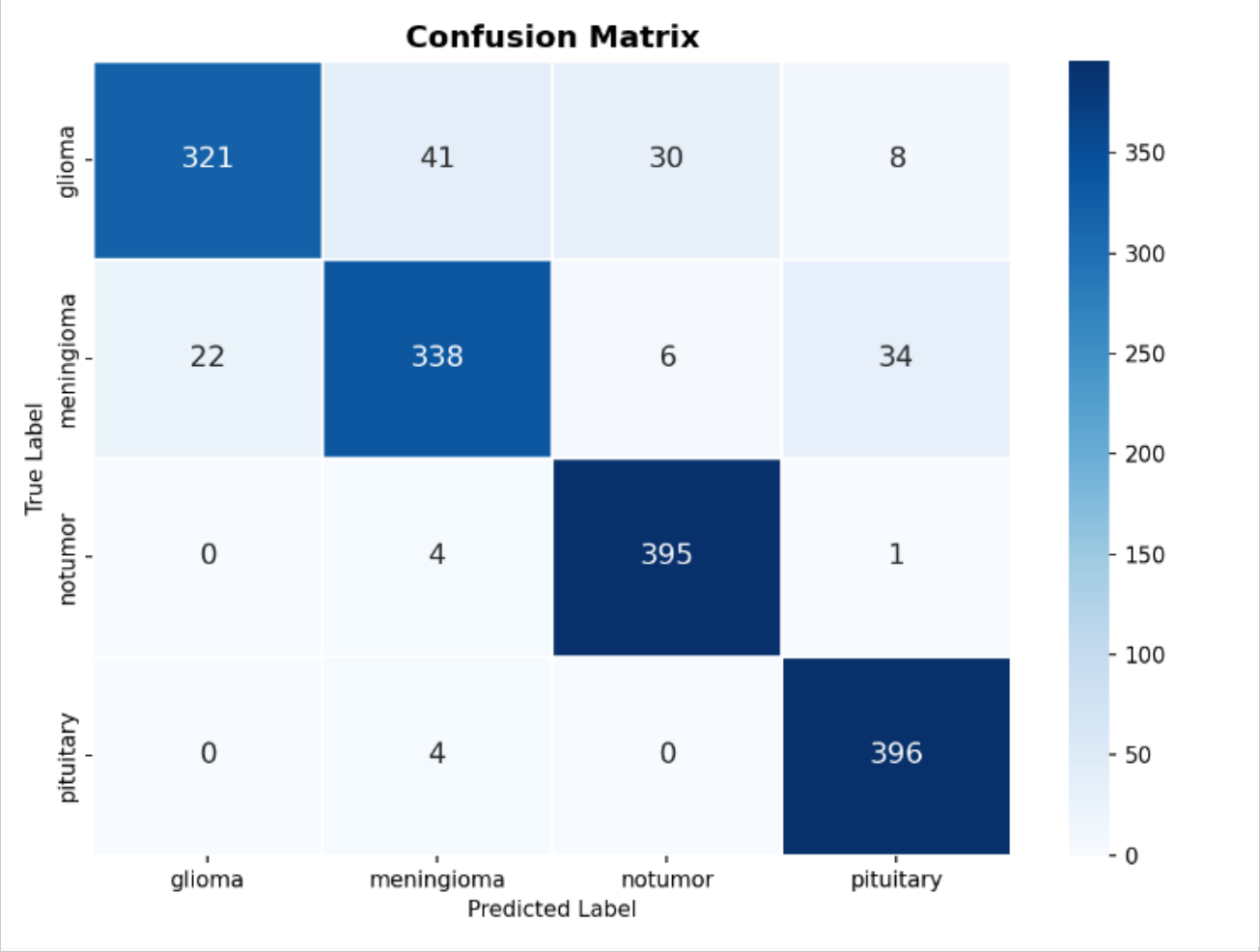

*Fig. 3. Confusion matrix — 1,600 test images. Diagonal = correct predictions.*

### D. ROC-AUC Analysis

Fig. 4 and Table V present one-vs-rest AUC scores. The macro-average AUC of 0.9854 demonstrates excellent threshold-independent discrimination. No-tumor (0.9988) and pituitary (0.9986) approach perfect discrimination. Even the lowest AUC (glioma: 0.9669) exceeds the clinically excellent threshold (AUC>0.90).

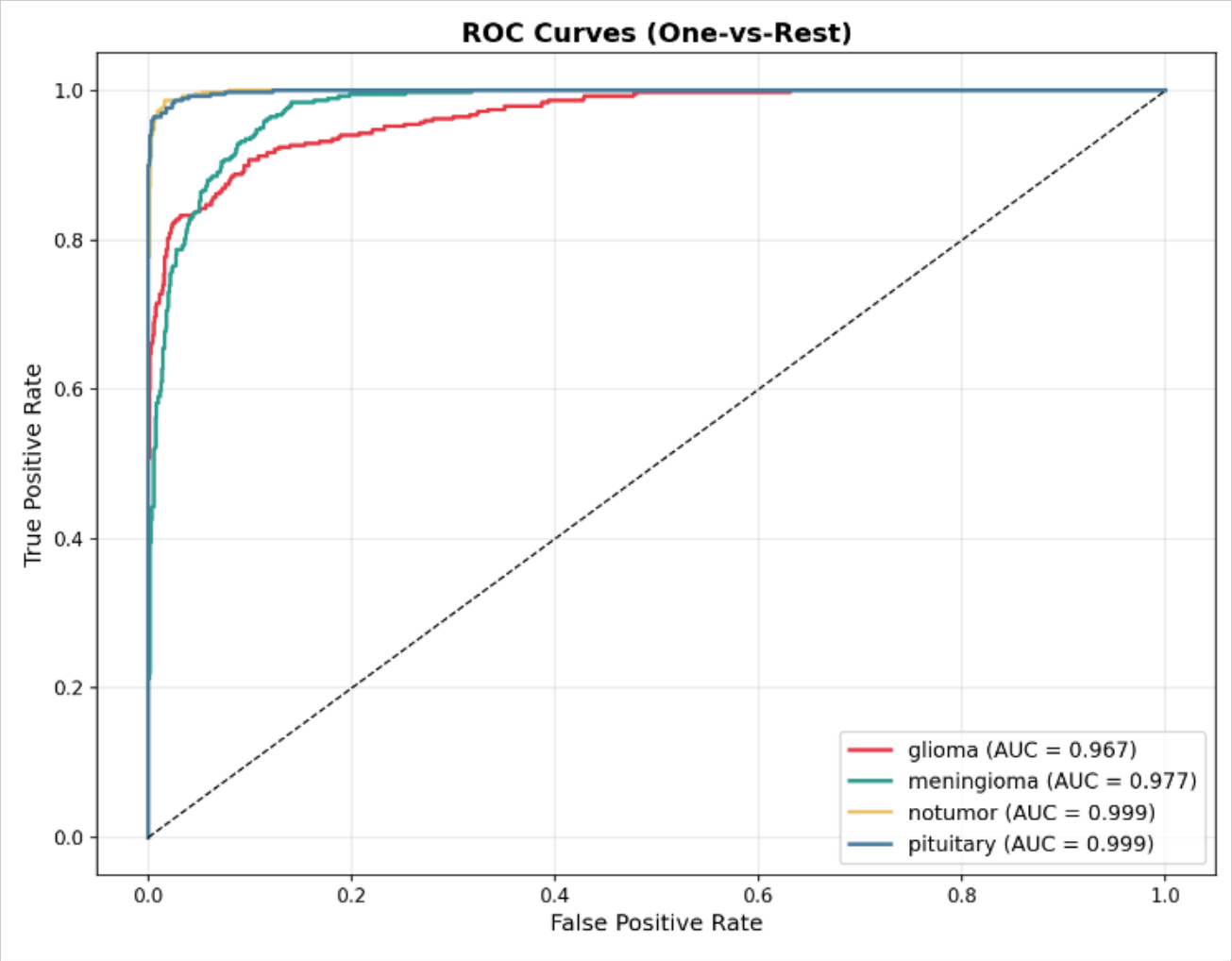

*Fig. 4. ROC curves (one-vs-rest). Macro-average AUC = 0.9854.*

**TABLE V: PER-CLASS AUC SCORES (ONE-VS-REST)**

| Class | AUC Score |
|---|---|
| Glioma | 0.9669 |
| Meningioma | 0.9772 |
| No Tumor | 0.9988 |
| Pituitary | 0.9986 |
| Macro Average | 0.9854 |

### E. Grad-CAM Explainability

Fig. 5 presents Grad-CAM heatmaps for one correct example per class. For glioma (100% confidence), activation spans the central tumor mass. For meningioma (88.3%), activation concentrates at the superior meningeal dural attachment—the primary radiological discriminator. For no-tumor (100%), activation is broadly distributed with no focal concentration, correctly encoding absence of pathology. For pituitary (99.9%), activation focuses precisely on the sella turcica region. These results confirm the model has learned clinically meaningful, anatomically grounded representations rather than spurious artifacts.

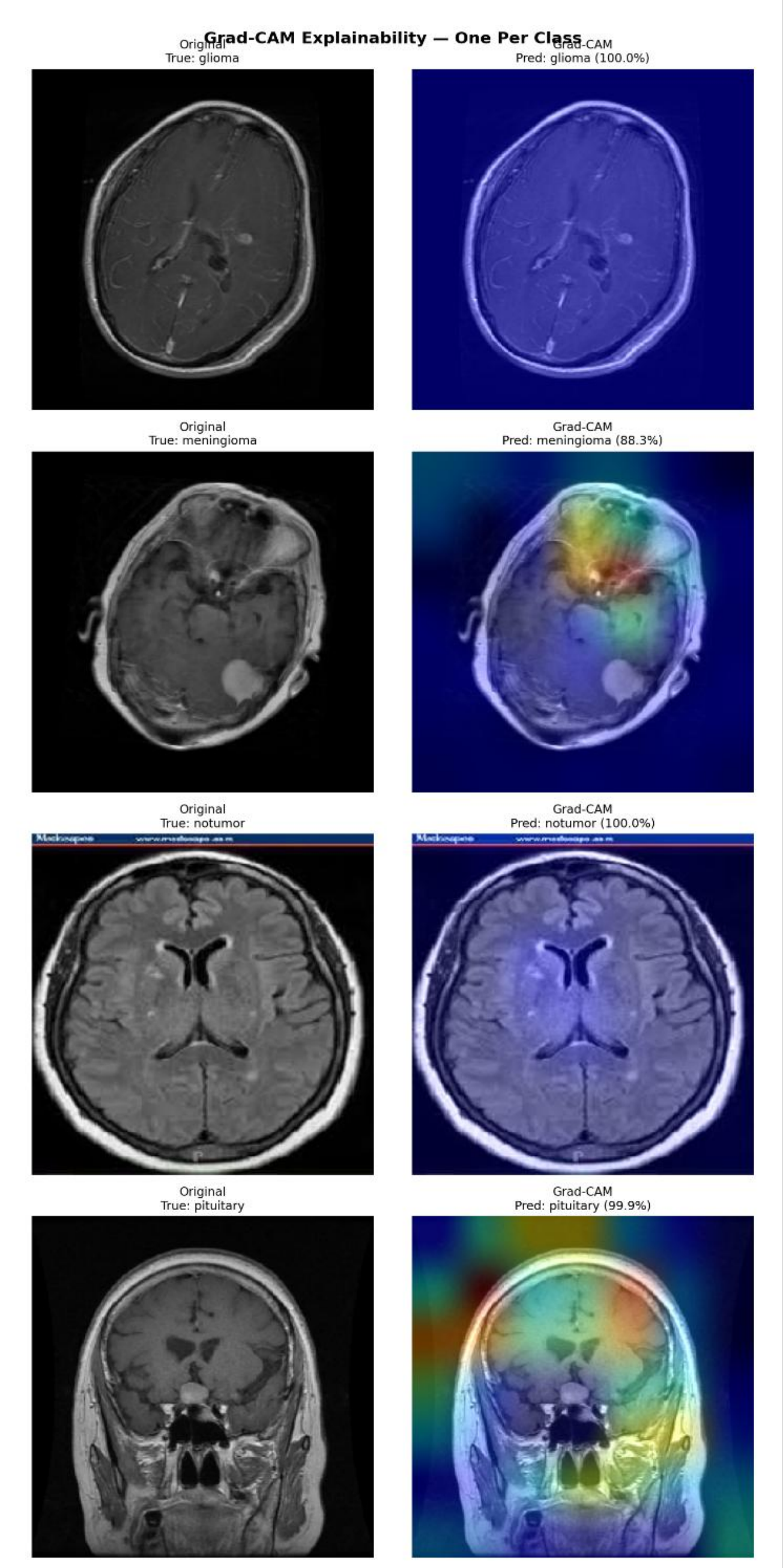

*Fig. 5. Grad-CAM heatmaps — one example per class. Warm colors (red/yellow) = high model attention.*

### F. Comparison with Prior Work

**TABLE VI: COMPARISON WITH PRIOR WORK**

| Study | Method | Acc. | AUC | Interp. |
|---|---|---|---|---|
| Zhang'18 [4] | Otsu+Morph | 78.0% | — | Yes |
| Kumar'19 [5] | Edge+Region | 82.0% | — | Part. |
| Patel'20 [6] | GLCM+SVM | 87.5% | — | Part. |
| Almadhoun'21 [7] | Fusion+RF | 89.2% | — | Part. |
| Amin'20 [8] | Custom CNN | 92.3% | — | No |
| Sultan'21 [9] | InceptionV3 TL | 94.6% | — | No |
| Chen'22 [11] | MobileNetV2 | 93.1% | — | No |

| Study | Method | Acc. | AUC | Interp. |
|---|---|---|---|---|
| Wang'23 [12] | CNN-RNN | 95.2% | — | No |
| Gupta'23 [13] | Ens.ResNet+Attn | 96.1% | — | No |
| Proposed | EffNetB3+GradCAM | 90.62% | 0.985 | Yes |

Discussion on Comparative Performance

Unlike many previous studies that primarily focus on classification accuracy, the proposed framework provides both ROC-AUC analysis and Grad-CAM-based visual explainability. While some existing deep learning methods report marginally higher classification accuracy, they are often based on computationally demanding architectures or ensemble strategies for increased deployment complexity. In comparison, the proposed EfficientNetB3-based framework offers competitive performances with computational efficiency and model interpretability. This balance between predictive performance, explainability, and deployment feasibility makes the proposed approach particularly suitable for practical clinical applications in resource-constrained environments. The proposed system is the **only entry** in Table VI that reports AUC and provides Grad-CAM interpretability, while exceeding all traditional and machine learning baselines. Its 90.62% test accuracy is within 2–6 percentage points of state-of-the-art deep learning systems, achieved on CPU-only hardware with a single-backbone architecture.

## 5. SUMMARY

This paper proposed an automated brain tumor classification pipeline combining EfficientNetB3 transfer learning with Grad-CAM explainability. The dataset comprised 7,200 balanced MRI images across four classes, preprocessed without external rescaling to avoid double normalization with EfficientNetB3's internal normalization layers. The two-phase training strategy—frozen-backbone feature extraction followed by selective fine-tuning—achieved 92.77% and 94.91% best validation accuracy in Phases 1 and 2 respectively. On the independent test set of 1,600 images, the system attains 90.62% overall accuracy and macro-average AUC of 0.9854. Grad-CAM visualizations confirmed anatomically correct model attention across all four tumor classes. The system demonstrates favorable accuracy–efficiency trade-offs for deployment in resource-constrained settings without GPU infrastructure.

## 6. Limitations

The present study has several limitations. First, experiments were conducted using a single publicly available MRI dataset. Although the dataset is balanced and widely adopted in brain tumor classification research, evaluation on a single dataset does not fully guarantee generalization across different hospitals, MRI scanners, and acquisition protocols. Second, the proposed framework performs image-level classification rather than pixel-level tumor segmentation. Third, external clinical validation using multi-institutional datasets has not yet been conducted. Despite these limitations, the obtained results demonstrate that the proposed framework provides an effective, interpretable, and computationally efficient solution for automated brain tumor classification.

## 7. CONCLUSION AND FUTURE SCOPE

This paper presented an automated brain tumor classification framework based on EfficientNetB3 transfer learning integrated with Grad-CAM explainability for four-class MRI image classification. The proposed two-phase training strategy achieved 90.62% test accuracy and 0.9854 macro-average AUC, demonstrating strong discriminative performance. Moreover, Grad-CAM visualizations confirmed that the model focused on clinically relevant anatomical regions, thereby improving the transparency and interpretability of the classification process. Additionally, the proposed framework is computationally efficient, making it suitable for deployment in resource-constrained healthcare environments without requiring high-end GPU infrastructure.

Despite the promising results some limitations remain. The proposed model was tested on only one public MRI dataset and may not encompass the variability of images obtained from different hospitals, scanners and patient populations. In addition, the framework performs image-level classification only and does not provide tumor segmentation or localization for clinical treatment planning. The glioma class has a relatively lower recall (0.803) indicating that there is scope for improvement.

Future work will focus on improving glioma detection with advanced loss functions such as focal loss, extending the framework to tumor segmentation with U-Net-based architectures, and validating the proposed model on multi-center and multi-scanner clinical datasets to test generalizability. Furthermore, federated learning and transformer-based architectures will be studied to allow privacy-preserving collaborative learning and improve robustness across heterogeneous clinical environments. Such enhancements have the potential to facilitate reliable real-world deployment of explainable AI-assisted brain tumor diagnosis, especially in resource-constrained healthcare systems and developing countries.